\documentclass[11pt,a4paper]{article}
\usepackage[latin1]{inputenc}
\usepackage[margin=1in]{geometry}
\usepackage{url}
\usepackage{hyperref}
\usepackage{amsmath,amsfonts,amsthm} 
\usepackage{wrapfig}
\usepackage{graphicx}
\usepackage{float}
\usepackage{fancyvrb}
\usepackage{subcaption}
\usepackage{comment}
\usepackage{enumitem}
\usepackage[toc,page]{appendix}
\usepackage{hyperref}
\usepackage[nameinlink,noabbrev]{cleveref}
\usepackage{cuted}
\usepackage{lineno}
\usepackage[ruled, vlined, linesnumbered]{algorithm2e}
\usepackage[noend]{algorithmic}
\usepackage[bottom]{footmisc}

\hypersetup{
    colorlinks=true,
    linkcolor=blue,
    filecolor=magenta,      
    urlcolor=cyan
}

\DeclareMathOperator*{\argmin}{argmin}

\title{Solving Visual Analogies using Neural Algorithmic Reasoning}
\author{
    Atharv Sonwane\footnote{corresponding author at f20181021@goa.bits-pilani.ac.in\newline \ \ \ \  Work done as part of an internship at TCS Research} \textsuperscript{\rm 1},
    Gautam Shroff \textsuperscript{\rm 2}, 
    Lovekesh Vig \textsuperscript{\rm 2}, \\
    Ashwin Srinivasan \textsuperscript{\rm 1}, 
    Tirtharaj Dash \textsuperscript{\rm 1} \\
    
    \textsuperscript{\rm 1} APPCAIR, BITS Pilani, K.K. Birla Goa Campus\\
    \textsuperscript{\rm 2} TCS Research, New Delhi\\
}

\date{}

\begin{document}

\maketitle

This document contains (1) the extended abstract for our work accepted at the AAAI-22 Student Abstract and Poster Program; (2) relevant supplementary material. In particular, the supplementary material describes details of the methodology adopted as well as additional results and observations. Note that while notation in the extended abstract has been simplified for brevity, the supplementary material makes use of much more extended formalism.

\tableofcontents 

\pagebreak

\section{Extended Abstract}

\begin{abstract}
We consider a class of visual analogical reasoning problems that involve
discovering the sequence of transformations by which pairs of input/output images are related, so
as to analogously transform future inputs. This program synthesis task can be easily solved 
via symbolic search. Using a variation of the `neural analogical reasoning'
approach of \cite{Velickovic2021NeuralAR}, we instead search for a sequence
of elementary neural network transformations that manipulate distributed representations
derived from a symbolic space, to which input images are directly encoded.
We evaluate the extent to which our `neural reasoning' approach generalizes for images with unseen shapes and positions.
\end{abstract}

\subsection{Introduction}

We consider a class of simple visual reasoning tasks specified by pairs of 
(input, output) images, all of which are related by the same unknown transformation procedure.
Given a new image, the task is to generate the correct output in an analogous
manner to the examples provided. We model this task as a program synthesis problem,
where the mapping between input and output is represented by a composition of elementary 
transformations. Building on previous program synthesis approaches which construct complex 
programs from simpler primitives, we employ a variation on the neural algorithmic reasoning
\cite{Velickovic2021NeuralAR} framework to replace elementary symbolic transformations
with equivalent neural networks, and examine if this leads to the generalisation which is the basis of 
analogical reasoning.

\begin{figure}[!htb]
    \centering
    \includegraphics[width=0.5\textwidth]{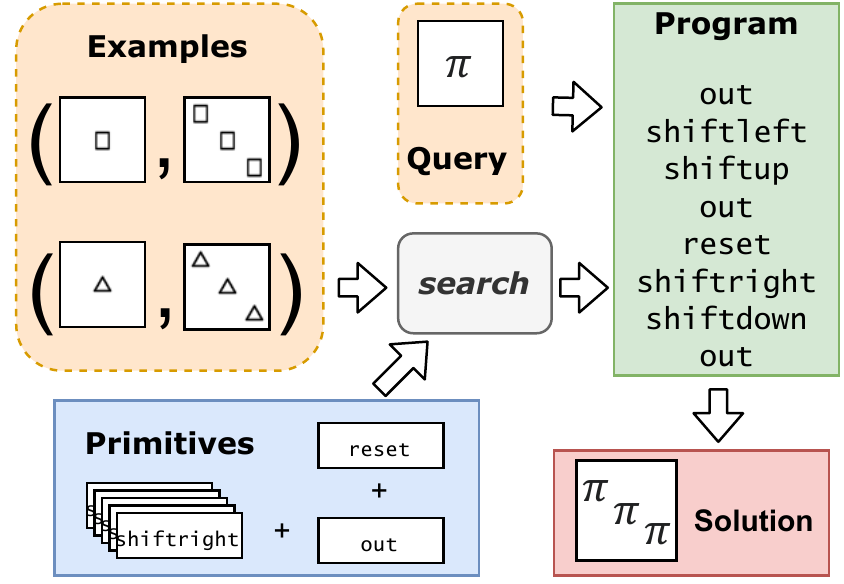}
    \caption{Overview}
    \label{fig:overview}
\end{figure}

We decouple representation learning 
from transform learning, first learning a latent representation that captures
key information from the input followed by learning transforms inside this 
representation space for which input-output examples have been given. 
Fig \ref{fig:overview} provides an overview of our approach wherein we search over possible combinations of primitives until one is found that satisfies all the example pairs. This solution is thus a neural algorithm,
i.e., a composition of elementary neural networks, rather than a symbolic program, corresponding to the \textit{unknown} analogical relation, which we can then apply to any query image to obtain an analogous result (output image).

\subsection{Methodology}

The images considered consist of simple shapes (from a 20 member set of alphabets and polygons) placed on a $3\times3$ grid while positional shifts and shape conversions make up the elementary transforms with which to compose our solutions. 

\paragraph{Representation Learning} 

Let a latent space be characterised by an encoder $E$ that maps natural inputs (e.g. images) to their latent representations (real-valued vectors). Since we are interested in spaces that are rich enough to 
represent a wide variety of shapes and sparse enough to represent concepts distinctly, we first construct a space based on symbolic descriptions of the input images containing information on the shapes and their positions in the 2D grid. For this, we train an autoencoder $(\text{Encoder: }E_{s}, \text{Decoder: }D_{s}$ on the multi-hot (Boolean) vector representations of symbolic descriptions by minimising the negative log-likelihood loss between the inputs and the reconstructed vectors. 

To be able to handle image inputs (denoted by $x$s), we train a CNN based encoder $E_{x}$ to fit to the outputs of $E_{s}$ (with frozen weights) by minimising the mean-squared-error over the encodings of images $E_{x}(x^{(i)})$ and encodings of the corresponding symbolic vectors $E_{s}(s^{(i)})$: $\text{MSE}\left(E_{x}(x^{(i)}),E_{s}(s^{(i)})\right)$.

To be able to handle shapes at test time that were not seen during training, we also include an additional shape-label called \textit{unseen}. We then train $E_{x}$ to map some set of shapes to the latent vectors generated by $E_{s}$ corresponding to the \textit{unseen} labels.

\paragraph{Transform Training}

We use single hidden layer MLPs as the neural transform networks ${\mathcal{T}_i}$s that manipulate the latent representations of the input images
where a transform refers to some spatial transformations on the original
space (images). 
For our experiments, we consider positional shifts in the ordinal directions (e.g. {\tt shift-right} moves each shape in the image one grid-position to the right) and shape conversions (e.g. {\tt to-square} converts each shape in the image to a square). 
We construct a separate transform network ${\cal T}_i$ for each such
spatial transformation, and learn its weights by minimising the negative log-likelihood between $D_{s}({\cal T}(E_{s}(s^{(i)}))$ and the multi-hot vector corresponding to the symbolic description of the transformed image, while keeping the encoder and decoder weights frozen.

\paragraph{Program Evaluation}

\begin{figure}[!htb]
    \centering
    \includegraphics[width=0.7\textwidth]{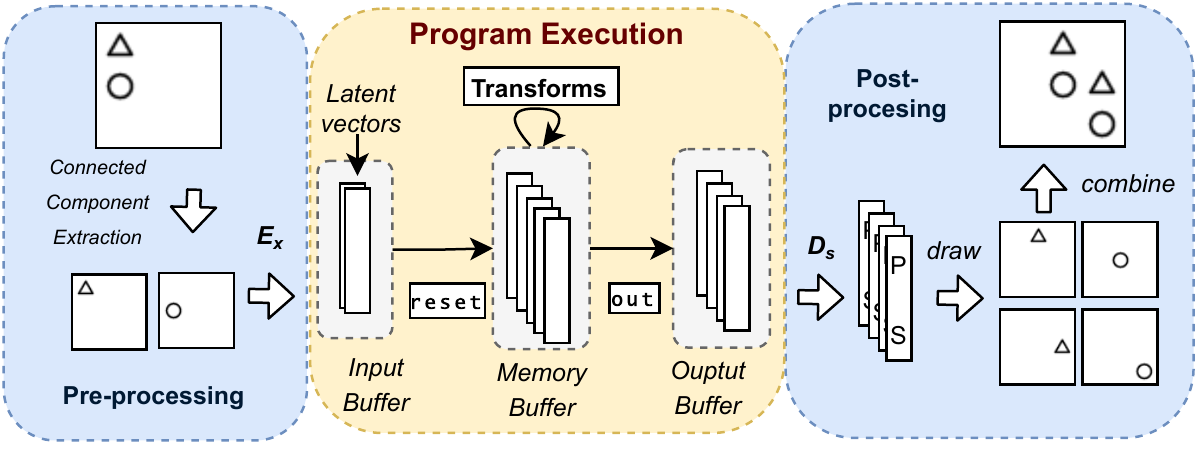}
    \caption{Pipeline}
    \label{fig:pipeline}
\end{figure}

We define primitives as transforms or special operations ({\tt reset} and {\tt out}); programs are sequences of such primitives. As shown in Fig.~\ref{fig:pipeline}, the input image is preprocessed to extract individual shapes by identifying the connected components. Each isolated image is converted into a latent vector using the encoder $E_{x}$ and stored in the input buffer. Algorithm~\ref{alg:exec} is applied and the resultant latent vectors in the output buffer are converted to their symbolic descriptions using $D_{s}$. 
The final image is constructed by drawing the corresponding shapes and combining them into a single image. (In the case of shapes marked as \textit{unseen} in the symbolic description, we keep track of the original shape from the input and use it to generate the output.)

\begin{algorithm}[!htb]
\caption{Program Execution}
\label{alg:exec}
\textbf{Input}: Set of latent vectors $z_k$s for input, program $P$\\
\textbf{Output}: Set of latent vectors for output

\begin{algorithmic}[1]
\STATE Let input = $\{z_k\}$, memory = $\emptyset$, output = $\emptyset$
\FOR{$i = 1 : \text{length}(P)$}
\IF {$P[i]$ = {\tt out}}              
\STATE output $\leftarrow$ output $\cup$ memory    
\ELSIF {$P[i]$ = {\tt reset}}             
\STATE memory $\leftarrow$ input  
\ELSE 
\STATE memory $\leftarrow \{P[i](z_k), \forall z_k \in \text{memory} \}$ 
\ENDIF
\ENDFOR
\STATE \textbf{return} output
\end{algorithmic}
\end{algorithm}

\paragraph{Searching for Solution} Given an (input, output) pair of images, we find the program that maps each input shape to the corresponding shape in the output by performing an exhaustive search over sequences of primitives. Various pruning measures are built into the search to make it tractable.

\subsection{Results and Discussion}

\begin{wrapfigure}{r}{0.45\textwidth}
    \centering
    \includegraphics[width=0.45\textwidth]{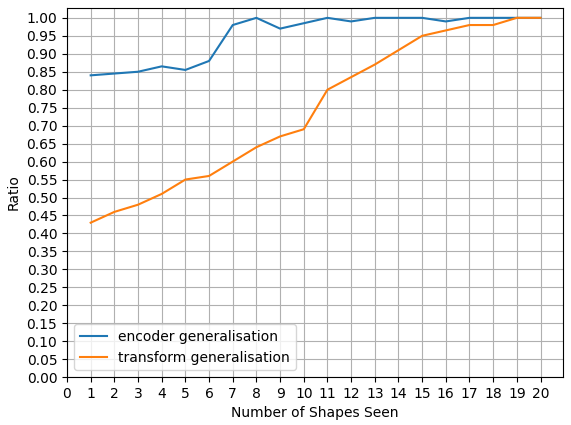}
    \caption{Generalisation Experiments}
    \label{fig:gen_exp}
\end{wrapfigure}

We evaluate our proposed system using ratio of (input, output) pairs for which a valid program is found. These pairs are generated from a dataset of 16743 programs (1000 programs for each length up to 20 with redundant programs removed). Fig.~\ref{fig:gen_exp} suggests that
our system is able to find the correct solutions for the complete dataset as well as all programs containing direct compositions of transforms up to length 5, demonstrating that conceptual integrity is maintained throughout successive transform application. This is not the case when using a latent space based on an autoencoder trained on images as opposed to multi-hot vectors, suggesting a well-structured embedding space is particularly important in this setting.

We evaluate how well transforms $\mathcal{T}_i$s and image encoder $E_{x}$ can generalise by withholding shapes during training and including them during evaluation. Note that when evaluating \textit{transforms} with varying number of seen shapes, the \textit{encoder} and \textit{decoder} have seen all 20 shapes during training. 
When evaluating the encoder, the \textit{unseen} label is used in addition to the labels corresponding to images seen during transform training; 
for this experiment, of the 20 shapes available, 4 have labels included in the symbolic descriptions while the rest are considered to have the label \textit{unseen}. From Fig.~\ref{fig:gen_exp}, we see that performance improves with more shapes seen during training (as expected). However, the system is able to generalise since it finds a valid programs a significant portion of the time even when observing very few shapes during training.
We observe similar results when evaluating generalisation to unseen grid-positions occupied by the shapes.



While our approach is limited by the use of classical algorithms for shape extraction as well as the mechanism for handling unseen shapes within a fixed symbolic code, we demonstrate within-domain generalisation capabilities (shapes seen during training the encoder but not transforms). Our approach, based on manipulating latent space representations, is, we submit, a step towards fully neural analogical reasoning with out-of-domain generalisation. 

\pagebreak

\begin{center}
    \Huge{Supplementary Materials}
\end{center}

\section{Detailed Methodology} \label{sec:method}

The ability to recognize and make analogies is often used as a measure of or to even test human intelligence \cite{hofstadter1995fluid}. We consider a simplified representative problem wherein the system need to identify and apply visual analogies from input output pairs of images related by some spatial transformation. 

We are given some $k$ (input, output) image pairs as examples of a spatial transformation. Here the images consist of simple shapes (from 20 member set of alphabets and polygons) placed on a $3\times3$ grid while the spatial transformation are composed of simpler elementary transformations consisting of grid position shifts (e.g. \verb|shift-right|) and shape conversions (e.g. \verb|to-circle|). We are also given a query image, on which we must apply the spatial transformation of which the preceding pairs are examples and generate the output. \\

We model this as a program synthesis problem where the analogous relationships (spatial transformations) are modeled as compositions of elementary primitives. We employ a variation on the neural algorithmic reasoning
\cite{Velickovic2021NeuralAR} framework to replace elementary symbolic transformations
with equivalent learned neural network transforms which manipulate latent representations of the input data. While \cite{Velickovic2021NeuralAR} propose learning a representation together with neural algorithms acting on abstract data, we decouple representation learning from transform learning, first learning a latent representation that captures key information from the input followed by learning transforms inside this 
representation space for which input-output examples have been given.

\subsection{Definitions} \label{sec:defs}
 
We consider two \textbf{domains}, $S$ denoting the space of symbolic descriptions of individual shapes in the image and $X$ denoting the space of images. Since images we are considering consist of some number of shapes placed on cells in an implicit grid, symbolic descriptions for each shape consist of the shape label and its position. We will be using the following variants of the above domains to describe images. $\mathbf{S} = \mathcal{P}(S)$ to denote the space of sets of symbolic descriptors (since an image may have more than one shape) and $\mathbf{X} = \mathcal{P}(X)$ where $\forall \mathbf{x} \in \mathbf{X}, |\mathbf{x}| = 1$ to denote the set of single element sets of images (this is necessary to maintain consistency in notation that an image is described by a  set). \\
 
Let the space of \textbf{board states} be defined as $B = \mathbf{S} \times \mathbf{X}$ so that a board state $b \in B$ is a two-tuple $b = (\{s_k\}, \{x\})$ composed of a set of symbolic descriptors  $s_k \in S\  \forall k$ for each shape in the given board and the image representation $x \in X$ of the board (given as a the only image in a set). \\
 
Let a \textbf{latent space} be defined as a 2-tuple $L(Z) = (E^Z, D^Z)$ where $E^Z : Z \rightarrow \mathcal{P}(R^n)$, $D_L : \mathcal{P}(R^n) \rightarrow Z$ such that $ \forall z \in Z, D^L(E^L(z)) = z$. We will consider two such latent spaces, $L(\mathbf{X})$ over the space of single element images sets $\mathbf{X}$ and $L(\mathbf{S})$ over the space of sets of symbolic descriptions $\mathbf{S}$. Note that the encoders and decoders work over sets since we are defining our board state as a set of objects. \\

The symbolic encoder and decoder will work element-wise over the input set to produce the output set. The image encoder will take the only element of the input set and generate as output a set of encoding for all shapes present in the image and the decoder will do the reverse.\\

Let a \textbf{program state} be defined as 3-tuple $PS(Z) = (I, M, O) \in Z \times Z \times Z $ where $I$, $M$, $O$ denote the input, memory and output set respectively and $Z$ can be any space such as $B$ or the internal representations from latent spaces such as $L(\mathbf{X})$ or $L(\mathbf{S})$.\\

We define \textbf{primitives} as mappings between programs states $T: PS(Z) \rightarrow PS(Z)$. \textbf{Transforms} are defined as a special type of primitive which only modify the contents of the memory $M$ and leave the input $I$ and output $O$ unchanged. Apart from transforms, the only two primitives currently considered are: \Verb|out| to denote copying all elements of memory $M$ to the output $O$ and \Verb|reset| to denote resetting of contents of memory $M$ with the input $I$. \\

A \textbf{program} $P$ is defined as a composition of primitives $P = T_1 \circ T_2 \circ \ldots \circ T_n$ such that $P(a) = T_{n}(...(T_{2}(T_{1}(a))))\ \forall a \in PS(Z)$. The procedure for \textit{applying} a program to an input board state is given in Algorithm  \ref{alg:app_program}. \\

An \textbf{example} $\lambda$ is defined as a 2-tuple $\lambda = (b_i, b_o)$ composed of the input and output board states. A \textbf{task} $\omega$ is defined as a 2-tuple $\omega = (\{\lambda_i\}, q)$  composed of a set of examples of a program $P$ and $q \in B$ being the query board state for which we need to be find the result on applying $P$. \\

Note that both the image and symbolic representations are included in board states so that algorithms can be written in a general manner independent of underlying domain. Which representation is to be used is selected according to the latent space in consideration. If this is known before hand (as in the case of final evaluation), during the creation of the tuples, the irrelevant representation can be kept empty since we know it will not be selected.

\subsection{Algorithm}

The formalised algorithm corresponding to the procedure demonstrated in \autoref{fig:overview} is shown below 

\begin{algorithm} 
\caption{General Algorithm for Solving Analogical Reasoning Tasks}
\label{alg:gen}
\SetAlgoLined
\KwIn{Search procedure: \Verb|search|, Max Depth: $d$, Latent Space: $L = (E^Z, D^Z)$ Transforms : $T^Z = \{T^Z_i\}$, Dataset of Tasks: $D_T = \{\omega_k\}$ where $\omega = (\{\lambda_i\}, q)$}
\KwOut{Solutions}
solutions $\leftarrow$ [\ ]; \\
\For{$(\{\lambda_i\}, q)_k \in D_T$}{
    P $\leftarrow$ \Verb|search|($d$, $L(Z)$, $T^Z$, $\{\lambda_i\}$) \\
    \eIf{P is not None}{
        g $\leftarrow$ \Verb|apply\_program|(P, $q$, $L(Z)$); \\
        solutions.add(g); \\
    }{
        solutions.add(None); \\
    } 
}
\textbf{return} solutions;
\end{algorithm}

\begin{algorithm} 
\caption{{\tt apply\_program}}
\label{alg:app_program}
\SetAlgoLined
\KwIn{Program: $P = [T_1, T_2, \ldots, T_n]$, Input Board State: $q = (\{s^q_k\}, \{x^q\})$, Latent Space: $L(Z) = (E^Z, D^Z)$}
\KwOut{Board State $g = (\{s^g_k\}, \{x^g\}) \in B$ obtained after applying $P$ to $q$}
encodings $ = \{e_j\}$ $\leftarrow$ $E^Z($\Verb|filter|$(q, L(Z))$ \\
\For{$e_j \in encodings$}{
    decodings $\leftarrow \emptyset$ ; \\
    memory $\leftarrow e_j$; \\ 
    \For{$i = 1 : \text{length}(P)$}{
        $T_k = P[i]$ \\
        \uIf{$T_k = $ \Verb|out|}{
            decodings $\leftarrow$ decodings  $\cup$ memory;
        } \uElseIf{$T_k = $ \Verb|reset|} {
            memory $\leftarrow e_j$;
        } \Else {
            memory $\leftarrow$ $T_k$(memory) // implied that $T_k$ is working on complete program state;
        }
    }
}
output $= (\{s^g_k\}, \{x^g\})$ $\leftarrow$ \Verb|unfilter|($D^Z$(decodings), $L(Z)$) \\
\textbf{return} output;
\end{algorithm}

\Verb|filter| is used to choose which (out of symbolic or image) representation to use out of a board state  and \Verb|unfilter| is used to construct a board state given only either symbolic or image representation.

\subsection{Latent Space}  \label{sec:latent}

\subsubsection{Encoder Decoder Architecture}

To obtain a latent space over some representation $Z$, we need to learn an Encoder and Decoder of the form $E^Z: Z \rightarrow \mathcal{P}(R^n)$ and $D^Z: \mathcal{P}(R^n) \rightarrow Z$ respectively. Since Z is either $\mathbf{S}$ or $\mathbf{X}$, elements of which are sets, the encoder and decoder need to work with sets as their inputs and outputs. In the symbolic case, we assume that the encoder and decoder neural networks act elementwise on the input sets to produce the output set. In the image setting, elements of $\mathbf{X}$ contain an image as their only element. There are three ways that this can be tackled - 

\begin{enumerate}
    \item Splitting up an image containing multiple shapes into multiple images containing single shapes using connected-component based image processing methods.
    \item Using an encoder architecture (such as RNN) which can take the image as input and output a set of latent embedding (each corresponding to one of the shapes in the image).
    \item Using only a single latent embedding for the complete image.
\end{enumerate}

Currently we are making use of the first method.

\subsubsection{Learning a Combined Representation Space}

\paragraph{Symbolic to Symbolic Setting} \label{sec:sym_latent_space} \ \\

\begin{figure}
    \centering
    \includegraphics[width=0.6\textwidth]{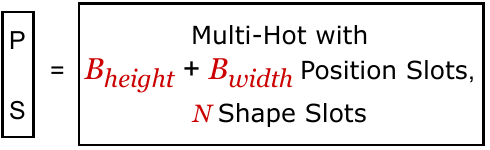}
    \caption{Multi-hot vector representation of symbolic descriptions}
    \label{fig:multihot}
\end{figure}

We obtain a latent space derived from an auto-encoder trained on the symbolic space $\mathbf{S}$ as $L(\mathbf{S}) = (E^\mathbf{S}_{\psi}, D^\mathbf{S}_{\phi})$ where $E^\mathbf{S}_{\psi} : \mathbf{S} \rightarrow \mathcal{P}(R^n$) and $D^S_{\phi}: \mathcal{P}(R^n) \rightarrow \mathbf{S}$ apply neural networks parameterised by $\psi$ and $\phi$ to the input set in element-wise manner to generate the output set. To be able to input the symbolic descriptions into the Neural Networks, we first convert them into a multi-hot vector of length $|N| + 2*S$ where $N$ is the set of shapes and $S$ is the size of the board. The multi-hot vector is made up of 3 concatenated one-hot vectors, one for denoting shape, one for x-coordinate and another for y-coordinate as seen in \autoref{fig:multihot}  \\

In this setting, $\psi$ and $\phi$ are obtained via gradient descent on a combination of negative-log-likelihood loss functions as follows - 

\begin{equation} \label{eq:ohe_loss}
    \mathcal{L}_{S}(g, t) = \Sigma_{k \in \{\text{shape}, \text{x}, \text{y}\}} - \log (\dfrac{e^{g^k_{\text{argmax}(t)}}}{\Sigma_i e^{g^k_i}})
\end{equation}

\begin{equation} \label{eq:transform_training}
    \psi, \phi = \argmin_{\psi, \phi} \Sigma_{s \in S} \mathcal{L}_{S}(\hat{s}, s) \text{  where  } \{\hat{s}\} = (D^S_{\phi}(E^S_{\psi}(\{s\}))
\end{equation}

\subsubsection{Image Setting}

There are two ways to learn the Latent Space in the case of the Image Setting. Either to base it off the Latent Space from the symbolic setting $L(\mathbf{S})$ or to obtain a completely new Latent Space for the Image setting in the same way as \autoref{sec:sym_latent_space} (by training an autoencoder followed by the transforms). \\

\textbf{Training Encoder to Target $L(S)$}\ \\

\begin{figure}
    \centering
    \includegraphics[width=0.6\textwidth]{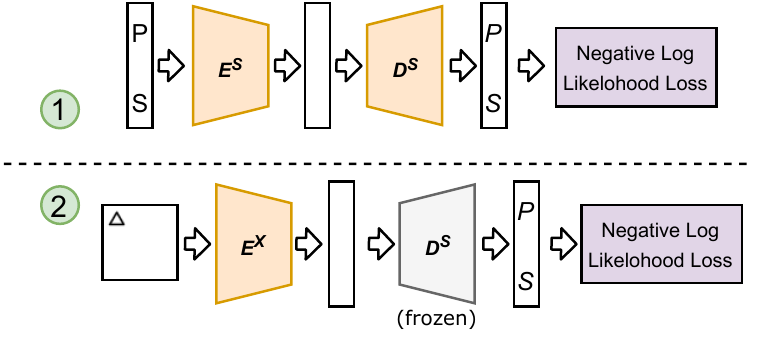}
    \caption{Procedure for obtaining image encoder using latent representation space based on symbolic descriptions}
    \label{fig:img_enc_target_training}
\end{figure}

We can make use of $L(\mathbf{S})$ when inputs and outputs are images by learning  $E^\mathbf{X}_{\psi} : \mathbf{X} \rightarrow \mathcal{P}(R^n)$ which applies a CNN based Neural Network to the only element image of the input set to fit to the outputs of the symbolic setting encoder $E^\mathbf{S}_{\psi}$ from \autoref{sec:sym_latent_space} as follows - 

\begin{equation} \label{eq:img_enc_sym_dec_training}
    \psi = \argmin_{\psi} \Sigma_{(\{s_k\}, \{x\}) \in B} \dfrac{1}{2}(E^\mathbf{X}_{\psi}(\mathit{\{x\}}) - E^\mathbf{S}_{\psi_\mathbf{S}}(\{s_k\}))^2
\end{equation}

This is demonstrated in \autoref{fig:img_enc_target_training}. Note the subscript to $\psi_\mathbf{S}$ to denote that these weights are from the symbolic space encoder and are distinct from the $\psi$ used to parameters $E^\mathbf{X}_\psi$. Using this image encoder and the decoder from $L(\mathbf{S}$), we construct a latent space $L^{S_{1/2}}(\mathbf{X}) = (E^\mathbf{X}_\psi, D^\mathbf{S}_\phi)$.\\

The decoder $D^\mathbf{X}_{\phi} : \mathcal{P}(R^n) \rightarrow \mathbf{X}$ in this setting can be trained in multiple ways. One being via a reconstruction loss while keeping encoder weights fixed  which gives the latent space $L^{\mathbf{S}_r}(\mathbf{X}) = (E^\mathbf{X}_{\psi}, D^\mathbf{X}_{\phi})$ where the parameters obtained are -

\begin{equation}
    \phi = \argmin_{\phi} \Sigma_{(\{s_k\}, \{x\}) \in B} \dfrac{1}{2}(D^\mathbf{X}_{\phi}(E^\mathbf{X}_{\psi}(\{x\})) - \{x\})^2
\end{equation}

Another being via the \textit{sandwich} loss approach which gives the Latent Space $L^{\mathbf{S}_s}(\mathbf{X}) = (E^\mathbf{X}_{\psi}, D^\mathbf{X}_{\phi})$ with parameters learnt as  -

\begin{equation}
    \phi = \argmin_{\phi} \Sigma_{(\{s_k\}, \{x\}) \in B} \dfrac{1}{2}(E^\mathbf{X}_{\psi}(D^\mathbf{X}_{\phi}(\tilde{h}_i)) - \{\tilde{h}_i\})^2
\end{equation}
\begin{equation}
    \{\tilde{h}_i\} = E^\mathbf{X}_{\psi}(\{x\})
\end{equation}

While in the proposed system, we reconstruct images from latent embeddings by first decoding to symbolic descriptions using $D^\mathbf{S}_{\phi})$ and then redraw those images according to these descriptions, the above approaches demonstrate that a end to end neural system is possible. 

\textbf{Obtaining a new Latent Space based on Image Setting}\ \\
Here we obtain a new Latent Space $L^a(\mathbf{X}) = (E^\mathbf{X}_{\psi}, D^\mathbf{X}_{\phi})$ by training an autoencoder as given below and then training transforms in a similar fashion to \autoref{eq:transform_training}

\begin{equation} \label{eq:img_autoencoder}
    \psi, \phi = \argmin_{\psi, \phi} \Sigma_{\mathbf{x} \in \mathbf{X}} \dfrac{1}{2}(D^\mathbf{X}_{\phi}(E^\mathbf{X}_{\psi}(\mathbf{x})) - \mathbf{x})^2
\end{equation}

\subsubsection{Handling Unseen Shapes} \label{sec:unseen_handling}

To be able to handle unseen shapes in the $L^{S_{1/2}}(\mathbf{X})$ setting, some additional measures need are taken. Since a multi-hot vector is used to represent symbolic descriptions of board elements while training the $E^\mathbf{S}$ and $D^\mathbf{S}$ in $L^{S_{1/2}}(\mathbf{X})$, it can only accommodate a finite set of shapes. To be able to handle any number of shapes, we add another shape class to the multi-hot vector for \textit{unseen} shapes.  This can be further generalised by removing the shape part from the the multi-hot completely. The issue then becomes, if there are multiple different shapes on the board, how do we tell them apart when the encoding contains no information about shape classes but only point positions. To handle this we keep track of which image corresponds to which latent embedding in the memory and output buffers. When trying to match if the multi-hot generated as output matches the targets, we make use of a distance metric between the image of the shapes corresponding to the generated and target multi-hots as described in \autoref{sec:search}.

\subsection{Transforms} \label{sec:transforms}

\textbf{Independent NNs for Transforms}\\

\begin{figure}
    \centering
    \includegraphics[width=0.8\textwidth]{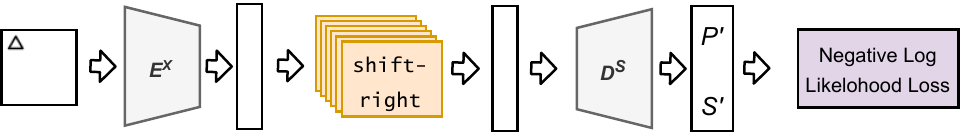}
    \caption{Procedure for training transforms}
    \label{fig:transform_training}
\end{figure}

As shown in \autoref{fig:transform_training}, we obtain a set of transforms $T^\mathbf{S}_{\theta}  = \{ T^\mathbf{S}_{i, \theta_i} \}$ over the latent space $L(\mathbf{S}) = (E^\mathbf{S}_{\psi}, D^\mathbf{S}_{\phi})$ where each $T^\mathbf{S}_{i, \theta_i}: \mathcal{P}(R^n) \rightarrow \mathcal{P}(R^n)$ applies a neural network parameterised by $\theta_i$ in an element-wise manner to input set to generate output set. The parameters $\theta_i$ are obtained via gradient descent on some loss function (either from \autoref{eq:ohe_loss} or MSE) while keeping weights of the encoder and decoder frozen.


\begin{equation}
    \theta_i = \argmin_{\theta} \Sigma_{b \in B} \\ \mathcal{L}(D^\mathbf{S}_{\phi}({\tt apply_program}([T^\mathbf{S}_{i, \theta}, {\tt out}], b, L(\mathbf{S}))),{\tt filter}(\bar{T}_i(b), L(\mathbf{S})))
\end{equation}

\textbf{Vectors for Transforms}\ \label{para:trans_vec} \\
An alternative to training separate sets of weights for each transforms is learning a vector $v_i \in R^k$ for each $T^\mathbf{S}_i$ along with a main transform network $T^\mathbf{S}_\theta : \mathcal{P}(R^n) \times R^k \rightarrow \mathcal{P}(R^n)$ which takes as input both the transform vector $v_i$ and the input set and generates the output set via element-wise application. \\

The {\tt apply\_program} procedure can be modified for this setting into {\tt apply\_program\_vectors} by taking $T^\mathbf{S}_\theta$ as an additional input $T$ and changing all instances of $T_k(memory)$ to $T(memory, v_k)$. The parameters $\theta$ and the transform vectors $\{v_i\}$ can be learnt via -


\begin{equation}
    \theta_i, \{v_i\} = \argmin_{\theta, \{v_i\}} \sum_{b \in B} \mathcal{L} (D^\mathbf{S}_{\phi}({\tt apply_program_vectors}([v_{i}, {\tt out}], T^\mathbf{S}_\theta, b, L(\mathbf{S}))), \  {\tt filter}(\bar{T}_i(b), L(\mathbf{S})))
\end{equation}

We use the \textit{Independent NNs for Transforms} method in our proposed approach but experimented with the above as an alternative.

\subsection{Search} \label{sec:search}

Algorithm \ref{alg:satisfies} is used to check whether a program using a certain latent space satisfies all examples in a given set. 

\begin{algorithm}[H]
\caption{satisfies}
\label{alg:satisfies}
\SetAlgoLined
\KwIn{List of Examples: $\{\lambda_i\}$ Program: $P$, Latent Space: $L(Z) = (E^Z, D^Z)$}
\KwOut{Whether $P$ satisfies all the examples in $B$}
 found $\leftarrow$ True; \\
    \For{$(b_i, b_o) \in \{\lambda_i\}$}{
        \If{$b_o$ $\neq$ \Verb|apply\_program|(P, $b_i$, $L(Z)$)}{
            found $\leftarrow$ False; \\
           \textbf{break}; \\
        }
    }
    \textbf{return} found;
\end{algorithm}

Note that the equality check here is non-trivial. Since the input output pairs can be given in any format (including natural representations such as images), there may not be a straightforward way to check if the generated output matches target output. In the case of images, we can make use of some continuous distance function and set some threshold below which two images can be considered equal. Options for the distance function includes -

\begin{itemize}
    \item Pixel based matching - two images are the same if their pixel values are exactly equal. While this works when using a symbolic system such as $L^{S_{1/2}}(\mathbf{X})$ described in \autoref{sec:unseen_handling} since output is constructed from shapes in the input, it will not work when images are generated by a decoder.
    \item Mean squared distance. This was tried in practice and seemed to work for simple shapes (squares, triangles etc...) but required careful tuning of the threshold each time the autoencoder was retrained.
    \item $||H - I||$ where $H$ is the homography \cite{Chum2005TheGE} between the detected and matching keypoints for two images which computes the amount of scaling, shearing and translation required to transform one image into another. This is a more appropriate metric since it explicitly takes into account the amount of spatial difference between individual shapes rather than just the difference in pixel values. (This is the method currently used for images).
    \item Learned distance function which could be trained in the style of a Siamese network.
\end{itemize}

We make use of the homography based option in our approach.

\subsubsection{Classical Search} \label{sec:classical_search}
The simplest search procedure that can be used in Algorithm \ref{alg:gen} to find a program that satisfies a given set of examples would be a exhaustive breadth first search upto some depth is given in Algorithm \ref{alg:naive_bfs}.

\begin{algorithm}[H] 
\caption{Limited Depth Exhaustive BFS}
\label{alg:naive_bfs}
\SetAlgoLined
\KwIn{Max Depth: $d$, Latent Space: $L(Z) = (E^Z, D^Z)$ Transforms: $T^Z = 
{T^Z_i}$, List of Examples: $\{\lambda_i\}$ }
\KwOut{Program that satisfies all given examples}
 q $\leftarrow$ LIFO(); \ q.push([\ ]) \\
 \While{True}{
    program $\leftarrow$ q.pop(); \\
  \If{length(program) = $d$}{
    \textbf{return} None;
   }
   \For{$t_i \in T^Z$} {
        new\_program $\leftarrow$ program + $t_i$; \\
        found $\leftarrow$ \Verb|satisfies|($\{\lambda_i\}$, new\_program, $L(Z)$); \\
        \If{found}{
            \textbf{return} new\_program;
        }
        q.push(new\_program); \\
   }
   
 }
\end{algorithm}

The naive implementation of BFS evaluates every possible sequence of transforms. With a branching factor of 8 (4 shift operations, 4 change shape operations), this implementation becomes intractable on our systems for searching for programs of length greater than 5. However, since we have both the final board as well as the initial board available during search, extensive pruning of the search tree is possible to reduce the effective branching factor significantly. The following measures were taken for pruning  -

\begin{itemize}
    \item If the current output buffer contains an element not present in target output, prune the current branch.
    \item Keeping track of the maximum number of matches with the target output so far by any branch. If a new branch has fewer matches than the maximum, it is pruned.
\end{itemize}

Along with this, we reduce the constant factor involved at each step by keeping track of the program state after partial execution. This creates a significant reduction in time taken since the majority of it is the neural network forward passes. This means that transforms are only applied when a new transform is added to the program. We do the same for output decoding - only applying output decoder whenever the \verb|out| primitive is called and keeping track of output buffer in subsequent nodes. The pruned version is described in Algorithm \ref{alg:pruned_bfs}

\begin{algorithm}[H]
\label{alg:pruned_bfs}
 \caption{Pruned BFS}
\SetAlgoLined
\KwIn{Max Depth: $d$, Latent Space: $L(Z) = (E^Z, D^Z)$ Transforms: $T^Z = 
{T^Z_i}$, List of Examples: $\{\lambda_i\}$ }
\KwOut{Program that satisfies all given examples}
 q $\leftarrow$ LIFO(); \ q.push([\ ]) \\
 \While{True}{
    program $\leftarrow$ q.pop(); \\
  \If{length(program) = $d$}{
    \textbf{return} None;
   }
   \For{$t_i \in T^Z$} {
        new\_program $\leftarrow$ program + $t_i$; \\
        found $\leftarrow$ \Verb|satisfies|($\{\lambda_i\}$, new\_program, $L(Z)$); \\
        \If{found}{
            \textbf{return} new\_program;
        }
        q.push(new\_program); \\
   }
   
 }
\end{algorithm}

\ \\



\subsubsection{Guided Search}

While an exhaustive search is guaranteed to find the solution program (if it exists), it makes no use of the given set of examples $\{\lambda_i\}$ during search apart from end checking and can become intractable for more complex settings (even with pruning). With our transforms being primarily based on spatial concepts, the input, output pair together contain information about the solution program which may be used to guide the search. \\

Formally, we want to model the distribution over programs given a set of examples as $P(T_1, T_2, ..., T_n | \{\lambda_i\})$. We can factorise this distribution over the complete sequence into one made up of distributions over next step - 

\begin{equation}
    P(T_1, T_2, ..., T_n | \{\lambda_i\}) = P(T_1 | \{\lambda_i\}) \Pi_{i=2}^{n}P(T_i | \{\lambda_i\}, T_1, ..., T_{i-1}) 
\end{equation}

This model can them be used to score program in a beam search as follows - \\

\begin{algorithm}[H]
 \caption{Guided Beam Search}
\SetAlgoLined
\KwIn{Max Depth: $d$, Latent Space: $L(Z) = (E^Z, D^Z)$ Transforms: $T^Z = 
\{T^Z_i\}$, List of Examples: $\{\lambda_i\}$, Program Likelihood Model: $P$ Beam width: $k$}
\KwOut{Program that satisfies list of examples}
beam $\leftarrow$ [$\phi$]; i $\leftarrow$ 0; \\
 \While{i < d}{
    \For{program in beam}{
       \For{$t$ in $T$} {
            new\_program $\leftarrow$ program + $t$; \\
            found $\leftarrow$ \Verb|satisfies|($\{\lambda_i\}$, new\_program, $L(Z)$); \\
            \If{found}{
                \textbf{return} new\_program;
            }
            beam.add(new\_program); \\
        }
    }
    beam $\leftarrow$ \Verb|select\_top\_k|(beam, $P$, $k$, $\{\lambda_i\}$);  i++;\\ 
}
\textbf{return} None;
\end{algorithm}

One way to formulate the \verb|select_top_k| procedure is by simply choosing programs with highest likelihoods. \\

\subsubsection{Rule Based Guidance}

As an initial step, we consider using hardcoded rules to build the distribution $P(T_1, T_2, ..., T_n | \{\lambda_i\})$. Since our input and output consists of a set of shapes, the simplest strategy is to score each program according to the number of matches between its output set and the target output. This is very similar to the pruning strategy used for making exhaustive search more tractable. Additionally, programs with shape conversion to a shape present in the target output are given higher scores.

\subsection{Data Generation} \label{sec:data}

Random examples can be generated by applying a randomly generated program to a randomly selected board. During generation of the program, constraints are applied (no consecutive shape conversion transforms, no consecutive \verb|out| or \verb|reset| operations, all programs must end with an \verb|out| operation, \verb|reset| operations must follow an \verb|out| operation) to minimise the generation of any redundant programs. When generating a set of random examples, it is also ensured that all examples generated are unique. \\

In this manner, we prepare two main datasets of programs. \verb|R20ALL| refers to a dataset of 16743 randomly generated programs containing 1000 programs of length 5 to length 20, 1 of length 1, 9 of length 2, 74 of length 3 and 659 of length 4 (total number of possible sequences under given constraints for those lengths). \verb|R20SHIFT| refers to a similar dataset as above but where programs contain only the shift operations along with \verb|out| and \verb|reset|. This contains a total of 15620 programs. During evaluation, we generate a dataset of examples by pairing randomly selected board with randomly selected programs from the above datasets.

\section{Further Results}

We consider the following factors with regards to our system in both the image to multi-hot ($L^{S_{1/2}}(\mathbf{X})$) and image-to-image ($L^a(\mathbf{X})$) settings -

\begin{itemize}
    \item How well are the transforms manipulating the latent space correctly?
    \item How well are the transforms generalising to unseen shapes and positions?
    \item How well is the latent space (encoder and decoder) generalising to unseen shapes and positions?
    \item Performance of the search procedures
\end{itemize}

\subsection{Evaluation of Latent Space}

\begin{figure}
    \centering
    \begin{subfigure}{\textwidth}
        \includegraphics[width=\textwidth, height=200pt]{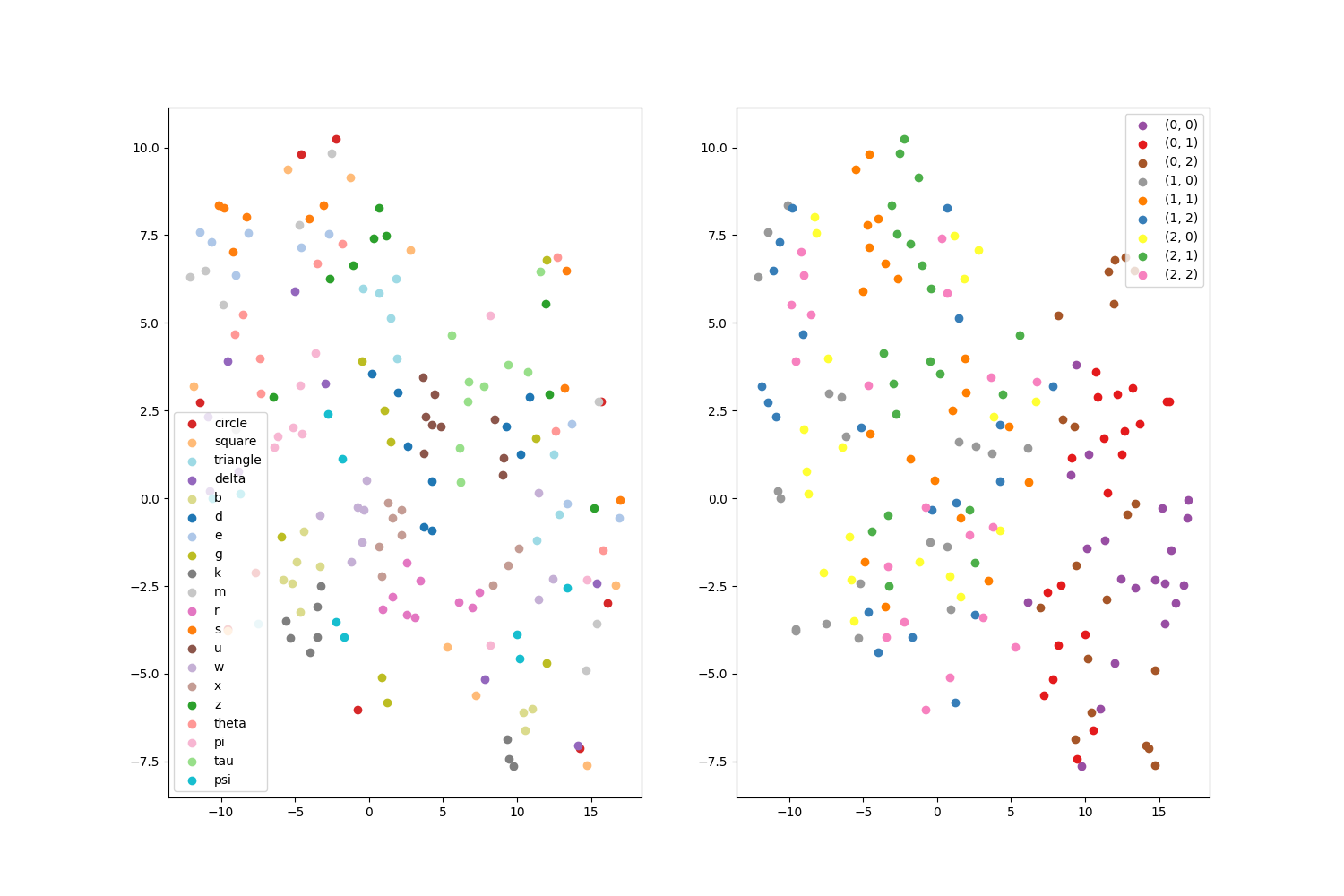}
        \subcaption{Image autoencoder}
    \end{subfigure}
    \begin{subfigure}{\textwidth}
        \includegraphics[width=\textwidth, height=200pt]{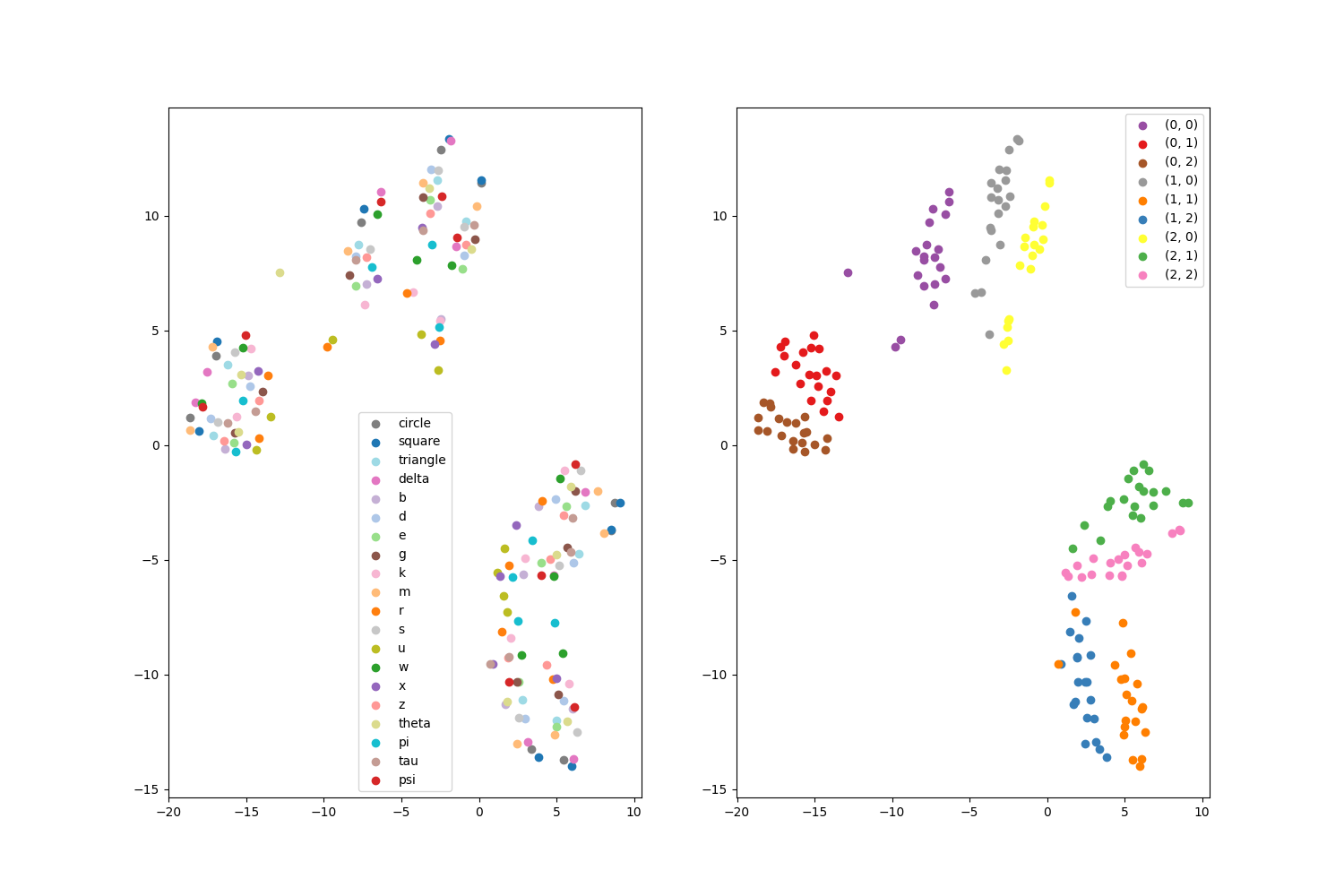}
        \subcaption{Image autoencoder with smaller CNN}
    \end{subfigure}
\end{figure}
\begin{figure}\ContinuedFloat
    \centering
    \begin{subfigure}{\textwidth}
        \includegraphics[width=\textwidth, height=200pt]{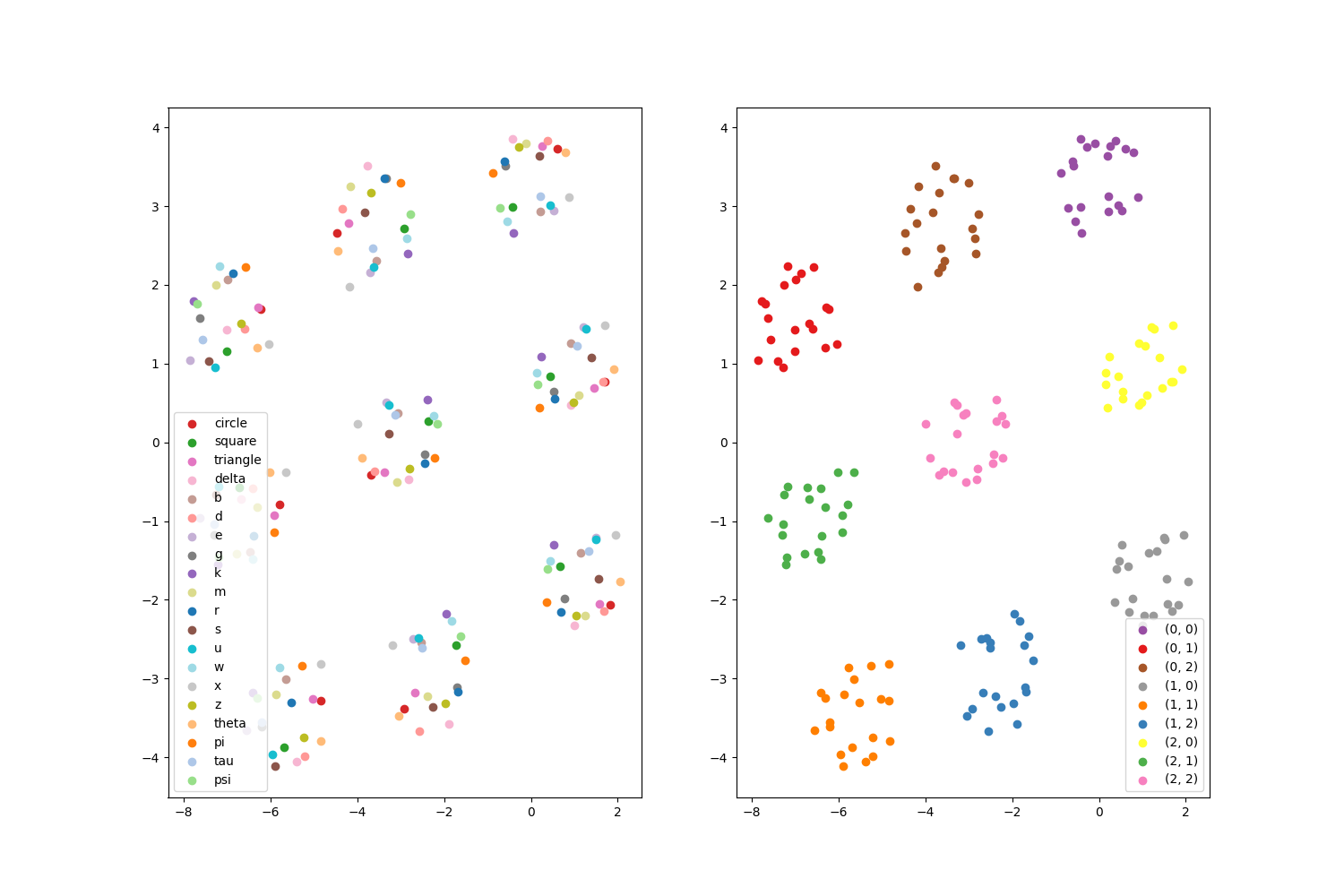}
        \subcaption{Ohe-hot autoencoder}
    \end{subfigure}
    \begin{subfigure}{\textwidth}
        \includegraphics[width=\textwidth, height=200pt]{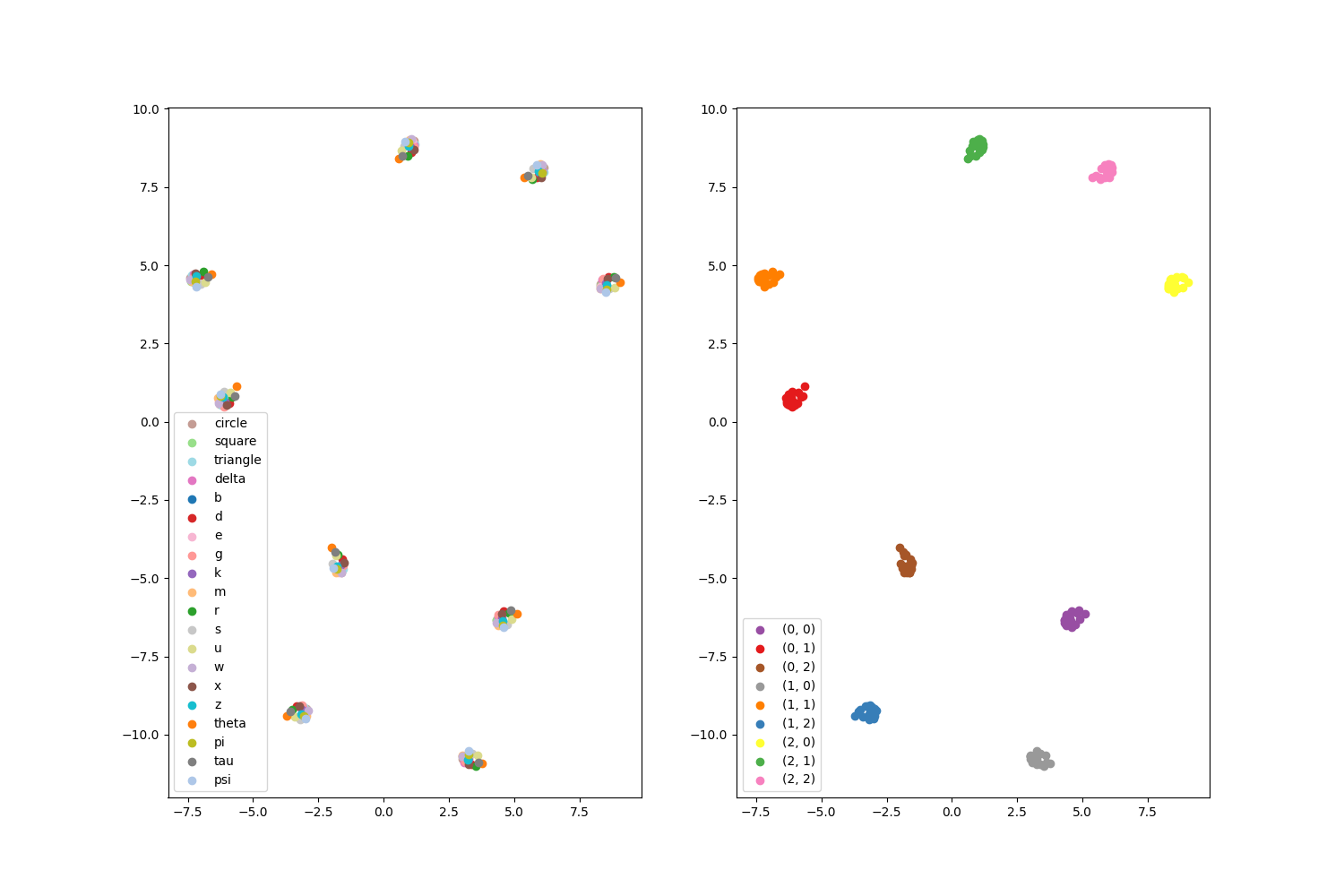}
        \subcaption{Ohe-hot autoencoder with mean-squared distance loss}
    \end{subfigure}
    \caption{TSNE Plots for Latent Embedding. Each point represents an embedding for a specific board element consisting of a single shape in a grid position on the board. The left hand plots have points coloured according to position and right hand plots according to grid position. }
    \label{fig:tsne}
\end{figure}

\begin{figure} 
    \centering
    \includegraphics[width=0.35\textwidth]{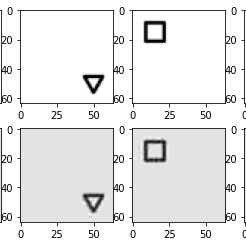}
    \caption{Reconstruction Examples using $L(S)$}
    \label{fig:recon_perf}
\end{figure}

In our system, the elements on which computation is to be done are latent embeddings in some vector space $R^k$ of the actual inputs (images or symbolic descriptions). Such a space is obtained by learning an autoencoder described in \autoref{sec:sym_latent_space}. It is important that this space is amenable to this computations. We utilise t-SNE to visualise and compare the distribution of latent embeddings in different cases in \autoref{fig:tsne}. \\

For the plot we can see that embeddings in the multi-hot space are much more structured than in the image space. This could be due to the difference in loss function (negative log-likelihood in the case of the multi-hots and mean-squared-distance in the case of the images). When using mean-squared distance loss for training the $L(S)$, we can see that the clusters for each position are much more tightly grouped, thus while it is clear on how to distinguish between positions, different shapes become indistinguishable. Another interesting phenomenon however is that while the space when using the smaller CNN is more structured, it also performs much worse in reconstruction only producing vague black shapes in the correct positions as compared to the almost perfect reconstructions of the larger autoencoder as seen in \autoref{fig:recon_perf}.

\subsection{Evaluation of Transform Correctness }

\subsubsection{Image to multi-hot setting}

To evaluate to what degree the transforms manipulate the latent space in the correct way, we apply Algorithm \ref{alg:naive_bfs} with transforms trained according to \autoref{eq:transform_training} on the dataset of examples (as described in \autoref{sec:data}). The hit-ratio is defined as the ratio of examples for which a program was found that gave the correct output from the given input. Note that the generated program may be different from the program used to create the example since multiple programs may be possible for a give input and output pair. \\

Evaluating transforms trained in $L(\mathbf{S})$ on programs with upto 6 consecutive shift transformations gives a perfect hit-ratio of 1.0. This indicates that these transforms can manipulate the latent embeddings accurately upto a length of 6. Similar evaluation was done on the transforms as vector (described in \autoref{sec:sym_latent_space}) setting. The results are displayed in \autoref{tab:acc_tab1}

\begin{table}[h]
\centering
\begin{tabular}{ |c| c c c c c| } 
 \hline
 Length of Sequence & 2  & 3  & 4  & 5 & 6  \\
  \hline
    Single Object Independent Transform & 1.0 & 1.0 & 1.0 & 1.0 & 1.0\\ 
 \hline
     Single Object Vector as Transform & 1.0 & 0.85 & 0.725 & 0.64 & 0.52 \\ 
 \hline
\end{tabular}
\caption{Accuracy of transforms trained in symbolic (multi-hot) setting with frozen latent space trained as autoencoder}
\label{tab:acc_tab1}
\end{table}

\subsubsection{Image to Image setting}

\begin{figure}
    \centering
    \includegraphics[width=0.35\textwidth]{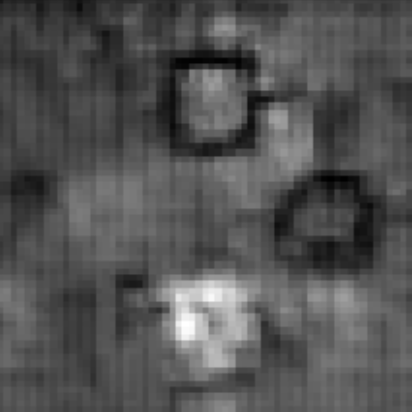}
    \caption{Reconstruction Example for image-to-image autoencoder}
    \label{fig:image_recon_perf}
\end{figure}

Training transforms in $L^a(\mathbf{X})$ with a loss similar to \autoref{eq:transform_training} resulted in poorly performing transforms. For example, when two consecutive \Verb|shift-left| transforms are applied to an latent embeddings of image with a single square in the middle row and left column and the result is decoder, we obtain the image in \autoref{fig:image_recon_perf}. It  is observed in general when using $L^a(\mathbf{X})$ that consecutive application of transforms leads to loss of conceptual information and the resultant image being incomprehensible. This highlights the importance of sparse and structured latent spaces as obtained  by training on symbolic descriptions.



\subsection{Evaluation of Latent Space Generalisation}

To test how well the latent space (encoder and decoder) generalises to unseen situations, we utilise a similar scheme as follows. Let us formally define it as a tuple $(M, N, R, T) \in \mathcal{P}(Q)^4 $ where $Q$ is the set of all shapes that can be present on a board, $M$ denotes the shapes present in the training data for the latent space (here $|M| + 1 + 6$ denotes the size of the multi-hot vector, here 6 is for the positional encoding and 1 is for a slot of \textit{unseen} shape), $N$ denotes the shapes present in the training data for transforms, $R$ denotes the shapes present in the training data for the encoder and $T$ denotes the shapes present in the examples on which the latent space is being evaluated. \\

\begin{figure}
    \centering
    \includegraphics[width=0.45\textwidth]{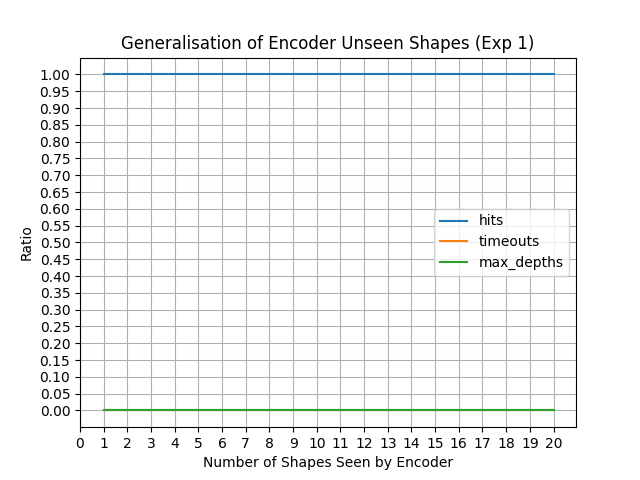}
    \includegraphics[width=0.45\textwidth]{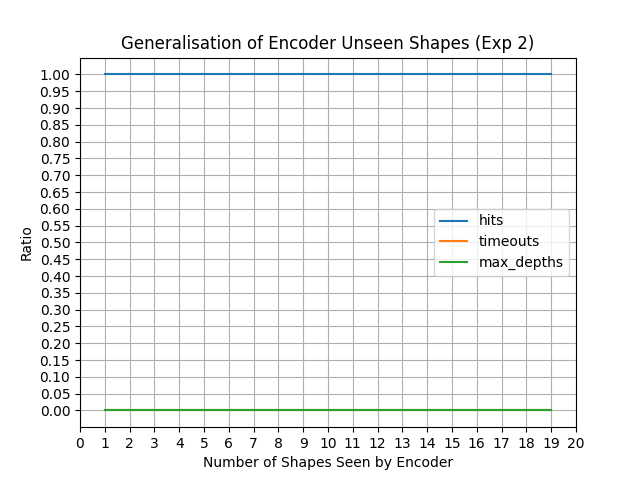}
    \includegraphics[width=0.45\textwidth]{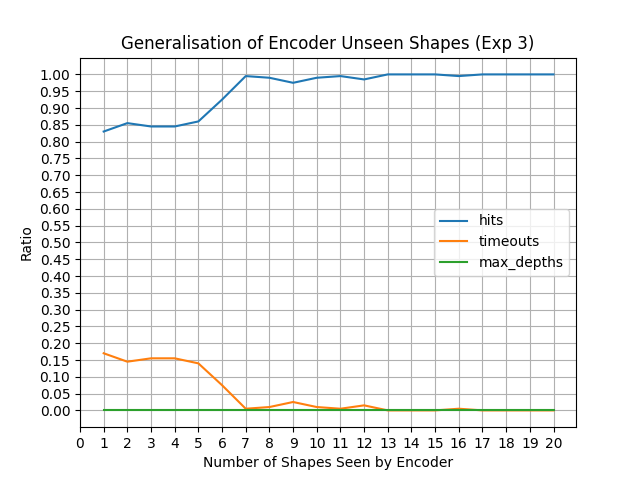}
    \includegraphics[width=0.45\textwidth]{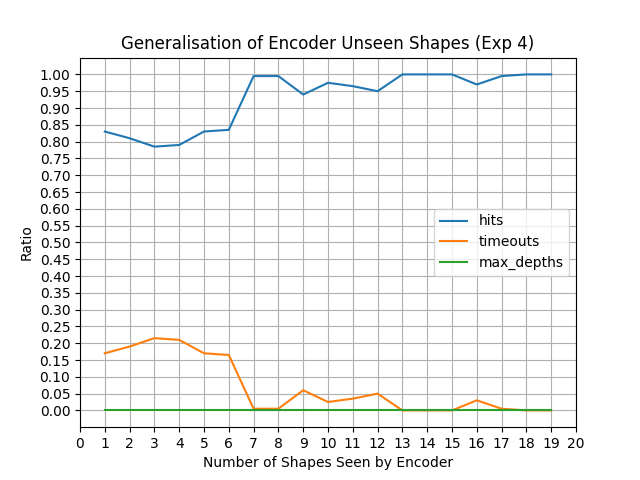}
    \caption{Generalisation of Latent Space to Unseen Shapes}
    \label{fig:sym_latent_unseen}
\end{figure}

We perform 4 experiments based on this evaluation scheme to test the generalisation capabilities of the latent space $L^{S_{1/2}}(\mathbf{X})$ where the decoder is from $L(\mathbf{S})$ and the encoder is a CNN based network trained to fit to the encoder from $E(\mathbf{S})$ as per \autoref{eq:img_enc_sym_dec_training}. The results are given in \autoref{fig:sym_latent_unseen}. Note that the total number of shapes possible $|Q| = 20$.  \\ - 

\begin{enumerate}
    \item For $M = N = \phi$ (all shapes considered unseen in the multi-hot) and $T = Q$ (all shapes included in test examples), we vary the shapes shown to the CNN encoder during training from $|R| = 1$ (just one shape being included in training data) $R = Q$ (all shapes included) and record the evaluation performance. 
    \item For $M = N = \phi$ (all shapes considered unseen in the multi-hot), we vary the shapes shown to the CNN encoder during training from $|R| = 1$ (just one shape being included in training data) $R = Q$ (all shapes included) and record the evaluation performance of with this encoder on $T = R'$ (test only including shapes not present in train for encoder). 
    \item For $M = N = \{\text{square}, \text{triangle}, \text{circle}, \text{delta}\}$ (four shapes + unseen in the multi-hot) and $T = Q$ (all shapes included in test examples), we vary the shapes shown to the CNN encoder during training from $|R| = 1$ (just one shape being included in training data) $R = Q$ (all shapes included) and record the evaluation performance. 
    \item For $M = N = \{\text{square}, \text{triangle}, \text{circle}, \text{delta}\}$ (four shapes + unseen in the multi-hot), we vary the shapes shown to the CNN encoder during training from $|R| = 1$ (just one shape being included in training data) $R = Q$ (all shapes included) and record the evaluation performance of with this encoder on $T = R'$ (test only including shapes not present in train for encoder). 
\end{enumerate}

\subsection{Evaluation of Transform Generalisation}

\begin{figure}
    \centering
    \includegraphics[width=0.5\textwidth]{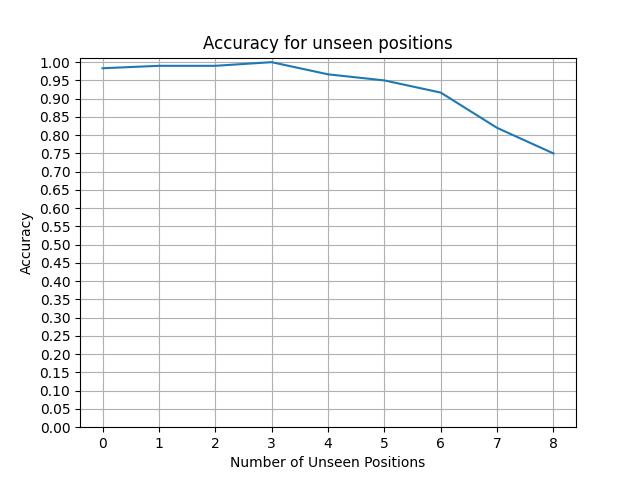}
    \includegraphics[width=0.5\textwidth]{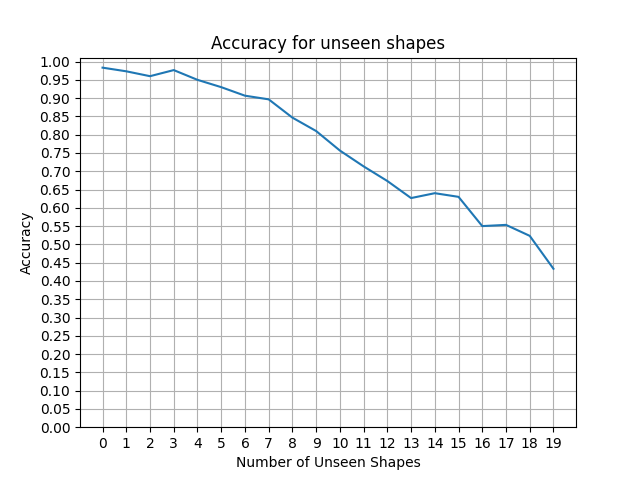}
    \caption{Generalisation of Transforms to Unseen Positions and Shapes}
    \label{fig:sym_transform_unseen}
\end{figure}

To test how well the learned transforms generalise to unseen situations in $L^{S_{1/2}}(\mathbf{X})$, we train the autoencoder with all the data, then train the transforms only on input, output pairs which have only a subset of shapes (or input positions). Then during testing of the transforms on a dataset of example programs, we include all shapes (or positions). This can be thought of as keeping $M = R = T = Q$ while $N$ is varied from $Q$ to $\phi$. If the transforms give similar performance to those trained with all the data, we can say that they, together with the latent space, are generalising well. The results for $L(\mathbf{S})$ are given in \autoref{fig:sym_transform_unseen}.\\

\subsection{Evaluation of Search Performance}

\begin{figure}
    \centering
    \includegraphics[width=0.55\textwidth]{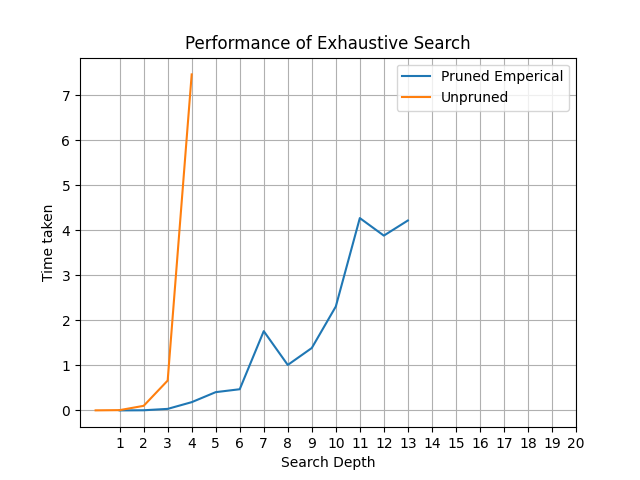}
    \caption{Search Performance}
    \label{fig:search_perf}
\end{figure}

The comparison between the running time performance of naive and pruned BFS (both discussed in \autoref{sec:search}) for program search is shown in \autoref{fig:search_perf}. We an clearly see the difference in performance and that pruning makes exhaustive search tractable when searching for longer programs.

Our simple rule based guidance was sufficient to make beam search with beam width of 500 perform well (80\%) upto search lengths of upto 15. Various other rules could be tried here however the high performance with the above simple rules gives sufficient evidence that a guidance system that understands the examples and the programs will be able perform well with beam search.

\pagebreak

\bibliographystyle{unsrt}
\bibliography{ref}

\end{document}